\documentclass[pdflatex,sn-mathphys-num]{sn-jnl}

\usepackage{graphicx}%
\usepackage{multirow}%
\usepackage{amsmath,amssymb,amsfonts}%
\usepackage{amsthm}%
\usepackage{mathrsfs}%
\usepackage[title]{appendix}%
\usepackage{xcolor}%
\usepackage{textcomp}%
\usepackage{manyfoot}%
\usepackage{booktabs}%
\usepackage{algorithm}%
\usepackage{algorithmicx}%
\usepackage{algpseudocode}%
\usepackage{listings}%

\usepackage{fontawesome5}
\usepackage{etoolbox}

\theoremstyle{thmstyleone}%
\theoremstyle{thmstyletwo}%

\theoremstyle{thmstylethree}%

\makeatletter
\def\blfootnotetext{\xdef\@thefnmark{}\@footnotetext}
\makeatother

\begin{document}

\title[ ]{Towards automated data analysis: A guided framework for LLM-based risk estimation}

\author*{\fnm{Panteleimon} \sur{Rodis}
\href{https://orcid.org/0000-0001-9169-8202} 
\orcid{}}

\abstract{
Large Language Models (LLMs) are increasingly integrated into critical decision-making pipelines, a trend that raises the demand for robust and automated data analysis. Current approaches to dataset risk analysis are limited to manual auditing methods which involve time-consuming and complex tasks, whereas fully automated analysis based on Artificial Intelligence (AI) suffers from hallucinations and issues stemming from AI alignment. To this end, this work proposes a framework for dataset risk estimation that integrates Generative AI under human guidance and supervision, aiming to set the foundations for a future automated risk analysis paradigm. Our approach utilizes LLMs to identify semantic and structural properties in database schemata, subsequently propose clustering techniques, generate the code for them and finally interpret the produced results. The human supervisor guides the model on the desired analysis and ensures process integrity and alignment with the task's objectives. A proof of concept is presented to demonstrate the feasibility of the framework's utility in producing meaningful results in risk assessment tasks.
}


\maketitle

\blfootnotetext{

\small 
\faEnvelope~rodis@uom.edu.gr,~~
\href{https://pantelisrodis.blogspot.com/}{\faLink}
\hfill May 2026
}

\section{Introduction}\label{introduction}

The impressive and rapidly evolving capabilities of foundation models have driven their integration into critical operational pipelines and complex decision-making processes. Concurrently, the exponential growth in data volume raises significant challenges in traditional data analysis practices and necessitates the development of automated solutions for risk assessment. 

There is a growing research interest in using LLMs  \cite{hong2025data, chugh2023intelligent}, or agents \cite{zhang2023data},to assist data analysis. But while the convergence of Generative AI and automated data analysis promises high value-added solutions \cite{sun2025survey, jansen2025leveraging}, current technology has not yet reached the maturity level required for fully autonomous operation \cite{nejjar2025llms}. 
The stochastic nature of LLMs introduces risks of hallucination and misalignment \cite{huang2025survey} that are unacceptable for crucial or sensitive data analysis scenarios. In essence, while it is valuable to leverage the semantic reasoning of LLMs, it is imperative to implement such methods within a Human-in-the-Loop architecture. 

To this end, we propose a guided framework for LLM-based risk analysis. Our approach leverages the advanced semantic analysis capabilities of LLMs and places the human as the supervisor of the process to filter undesired operations, ensure alignment with the task's objectives and possibly debug some of the generated code. The objective of this work is to develop a framework that will produce full data analysis reports with the highest possible degree of automation, rather than just some chat-based analysis assistant. This methodology constitutes the contribution of this work.

The proposed framework consists of four sequential stages, each comprised of distinct processes, rather than just a single, zero shot, LLM-based procedure that would be prompted to analyze the whole dataset. In this design, the Human-in-the-Loop can supervise the integrity of each stage and assess the intermediate results before moving to the next stage.

In the next section, we provide an overview of the proposed framework, followed by a discussion on the critical processes it utilizes in Section~\ref{description}. In Section~\ref{poc}, we present a proof of concept of the framework in the risk estimation of non-technical losses on power grids and finally Section~\ref{discussion} concludes the paper with a discussion on the potential and limitations of using agents and LLMs in risk analysis.

\section{Overview}
\label{overview}

\begin{figure}[htbp]
  \centering
  \includegraphics[width=0.85\columnwidth]{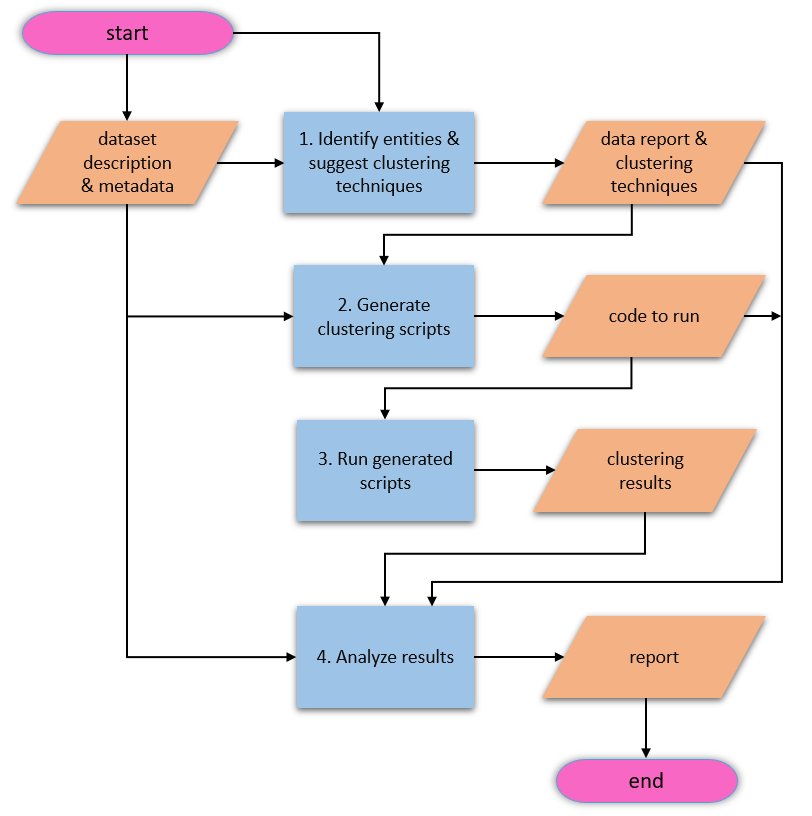}
  \caption{Flowchart of the proposed framework.}
  \label{fig:flowchart}
\end{figure}

In every stage of the procedure we initiate a new session with the LLM. We need to evaluate and possibly modify the output of each process before moving to the next, therefore an independent session is required in each stage to avoid the model being affected by context contamination caused by undesired intermediate data and failed past prompts. The flowchart that describes the overall procedure is illustrated in fig.~\ref{fig:flowchart}. The four main stages and their processes are defined as follows:

\begin{enumerate}
    \item The model is provided with a description of the dataset and the relevant metadata. It is then prompted to identify entities and relationships and subsequently to suggest clustering techniques. The model generates a structured description of the dataset, an essential step for the subsequent stages especially if the input is unstructured. LLMs exhibit higher computational efficiency when processing structured information data \cite{elnashar2025prompt}.
    
    Where considered necessary, additional functionality can be integrated to assess the quality of the dataset, either by prompting the model to generate processes for data quality evaluation that the user will execute or by directly accessing the dataset within an agentic AI implementation. Data quality evaluation, especially in cases of problematic datasets, can improve the efficiency of the subsequent stages.

    \item The generated data reports and the descriptions of the suggested techniques, along with the dataset description and metadata, are fed to the model which is prompted to generate scripts, e.g. Python, that implement the clustering techniques.
    
    It is case dependent whether a parameter optimization process will be integrated in this stage of the framework. If required, an interactive procedure including the Human-in-the-Loop has to be implemented to compute the optimal parameter values; this functionality was not required for the proof of concept presented in this work.

    \item Next, the generated code is executed, by the human supervisor or an agent, and the clustering results are stored.

    \item At the last stage, the model is prompted to analyze the produced results. It is provided with the clustering results alongside the dataset description, the data reports and the clustering techniques descriptions and code. It is then prompted to produce a conclusive report substantiating the results and generate a rating report for the audited entities.

    The model can either access the output of each algorithm and synthesize the final report, or in case of large intermediate outputs, it can generate additional scripts to analyze the content of the files and eventually produce the report.
\end{enumerate}

\section{Framework description}
\label{description}

\subsection{Entities and relationships identification}
\label{er}

The primary advantage in the utilization of LLMs for dataset analysis is their ability to bridge the gap between data structure and human interpretation. Traditional algorithms for semantic schema matching typically rely strictly on explicit constraints, like foreign keys or naming conventions, while LLMs treat a database schema as a linguistic corpus for interpretation. This shift allows for the identification of entities and relationships even within databases where explicit constraints are missing or obfuscated. Furthermore, LLMs can leverage information from diverse sources, even unstructured in natural language, to generate structured documentation \cite{brach2025effectiveness}, thereby augmenting database analysis in our case. Next, we detail the specific capabilities that facilitate this advanced LLM-based analysis.

\textbf{Schema Item Grounding.} The process incorporates a Schema Item Grounding mechanism to map abstract schema symbols to concrete real-world concepts. Leveraging table definitions and metadata, the model does not merely parse the data types but rather grounds abstract identifiers into semantic entities. Recent work \cite{han2025leveraging} demonstrates that utilizing LLM-generated schema descriptions significantly enhances the system's ability to understand underlying data distribution. The transformation of abstract metadata into semantically rich descriptions enables the effective conversion of the relationship detection task from a mathematical graph problem into a semantic compatibility problem. This approach aligns with findings presented in \cite{buss2025towards} that highlight the efficacy of generative models in reconstructing structural properties from flat schema definitions.

\textbf{Resilience to Non-Standard Nomenclature.} A critical challenge in automated risk analysis is the prevalence of non-standard naming conventions in certain databases. Traditional heuristic matching often fails when database field names are ambiguous or abbreviated. However, our framework leverages the LLM's resilience to such obfuscation. As shown in \cite{luoma2025snails}, LLMs can infer the semantic role of columns despite poor naming conventions by analyzing contextual information, data types and implicit usage patterns. This inherent capability enables the robustness of our framework even when dealing with datasets with high entropy in their naming schemata.

\textbf{Semantic and Logical Inference.} The identification of complex relationships relies on advanced logical reasoning. The model employs inductive and abductive inference to deduce implicit connections between entities, rather than simply matching explicitly defined patterns. As discussed in \cite{zheng2025logidynamics}, LLMs exhibit dynamic logical reasoning capabilities, termed LogiDynamics. These enable them to infer relationships based on partial evidence, even in the absence of explicit foreign key definitions, by inductively reasoning about the domain context. This reasoning is utilized by our framework to reconstruct a representation of the data topology, serving as the necessary foundation for the subsequent clustering and risk analysis stages.

\subsection{Clustering technique suggestion}
\label{suggestion}

The integration of LLMs in algorithm design and selection process constitutes a field of intense and accelerating research \cite{liu2024systematic}, yielding highly promising results. 
Leveraging their advanced analytical and discriminative capabilities, LLMs effectively bridge the gap between requirements description and concrete technical solutions by rapidly mapping prompt parameters and data descriptions, to relevant peer reviewed papers and established methodologies. Finally, they manage to suggest the most appropriate clustering algorithm for the task at hand. We identify two main features for this advanced  functionality.

\textbf{Extensive training.} LLMs are pre-trained on vast corpora of academic and technical literature and as such they possess a comprehensive overview of the current state-of-the-art.

\textbf{Feature based algorithm selection.} According to \cite{ wu2023large}, LLMs succeed in algorithm selection as they do not take under consideration problem features alone but inclusively evaluate algorithm features as well.

In this context, it is necessary to delineate the scope and expectations of the LLM-based method suggestion process. The model is actually an advanced recommendation engine of methods drawn from the literature, equipped with synthetic abilities of efficiently combining existing algorithms and best practices. Despite its ability to excel at optimizing known methods and applications, currently it is not expected to synthesize novel or fundamentally innovative algorithms.

\subsection{Code generation}
\label{code}

Leveraging LLMs for code generations, termed \textit{Vibe Coding}, seems to be a ubiquitous practice these days. The technical foundation for this shift lies in the inherent capabilities of LLMs for code analysis and generation. These models are trained on large codebases \cite{chen2021evaluating}, where code analysis is approached as a linguistic task \cite{yang2024if}. They exhibit advanced efficiency in code generation \cite{yang2024if} and program synthesis \cite{austin2021program}, \cite{allal2023santacoder}.

Despite these advances, code generation has not yet reached the expected maturity that would couple efficiency and trustworthiness \cite{liu2023your}.
There are two well studied reasons for this:

\textbf{Hallucination.} It refers to the phenomenon where an LLM produces content inconsistent with the prompt or any given data and fails to produce the expected output. This problem is sometimes due to incomplete training, while in many cases the stochastic nature of the model fails to estimate the correct probability distribution over the available choices. In code generation the model fails to estimate the impact of the produced code \cite{zhang2025llm}.

\textbf{Alignment issues.} This is an inherent problem in AI as we cannot take for granted that we will get what we asked for. There are many reasons for misalignment between inputted requests and failed output, which include unclear and ambiguous prompts by the user and misinterpretation of the prompts by the model, both result in undesired outputs \cite{gabriel2020artificial}.

These admissions necessitate the Human-in-the-Loop, often accompanied by iterative improvement. The workflow can involve a cyclical process of generation, execution, error assessment and prompt refinement. 

\subsection{Results analysis}
\label{reports}

To provide insights on the clustering results, we leverage the reasoning capabilities of the LLM. The model is provided with the dataset description, the clustering techniques and the code, along with the output files of the algorithms and is assigned to produce reports analyzing the results.

In an agentic AI implementation, the agents directly access the files and apply some report generation method. In this design, we rely on the reasoning and acting abilities of agentic frameworks \cite{yao2022react}.

Conversely, in case the model cannot process the files due to high data volume or privacy constraints, it generates a script that the user will run to produce the final analytical report. From this perspective, this stage is essentially a repetition of the first three stages of the procedure. The model assesses the structure and content of the algorithm outputs, identifies the entities they contain and their utility and subsequently suggests report generation methods that will be executed by the human supervisor.

A key difference from the first stage is that the files generated by the algorithms are by definition structured and of known rationale, while the initial dataset may be unstructured and lack proper documentation. As such, suggesting a report-generating method is essentially an easier task for the model than investigating clustering techniques. Consequently, running the generated code is expected to be a trivial task.

\section{Proof of concept: Risk estimation for non-technical losses in power grids}
\label{poc}

The proof of concept presented herein addresses non-technical losses risk estimation within the power grid in Greece. The objective is to assess the risk of individual accounts participating in non-technical loss activities, commonly referred to as electricity theft. 
Data analysis in this domain presents significant challenges due to the inherent irregularity and sparsity of the available datasets \cite{papadimitriou2017non}. Subscriber accounts exhibit heterogeneous characteristics in terms of temporal duration and the diversity of their usage features. Furthermore, data collection processes face significant challenges resulting in highly sparse and irregular datasets.

The dataset we used was provided by Hellenic Electricity Distribution Network Operator S.A. (HEDNO) during the 2023 Datathon \cite{HEDNO2023}.
It is a real-world dataset and presents a substantial pattern recognition challenge, characterized by irregular data collection patterns along with extensive inherent divergence and sparsity.

This dataset includes anonymized data of 1,234,509 consumer accounts, 9,201,395 consumption measurements and 3,842 cases where technical inspections have verified non-technical losses. Installations were located in an urban area in Greece and data was collected from early 2018 to late 2022. The absence of accounts confirmed not to exhibit in non-technical losses is an inherent characteristic of this problem domain and increases the complexity of the problem.
The provided data included provider information for each account, installation characteristics, subscriber request records, power supply technical characteristics and confirmed non-technical losses reports.

For the experimental evaluation, we employed Gemini 3.0 Pro as a state-of-the-art LLM. The evaluation was conducted on a workstation equipped with 15 GB of system RAM and an NVIDIA Tesla T4 GPU of 16 GB VRAM. Each process was efficiently executed with a runtime of three minutes or less.

In the first stage of the procedure, the model was provided with the original file describing the dataset, as it was distributed to the participants of the Datathon in Greek, containing both structured data description and descriptive text. Then we prompted the model as follows:

\vspace{10pt}
\fbox{
\parbox{0.9\textwidth}{%
\texttt{
PROMPT: In this file you are given the description of a dataset. 
Your task is to \\
"1. Identify possible relationships between the tables (e.g., foreign keys, shared fields)." \\
"2. Suggest clustering techniques that could be used to group similar data samples across these tables. The clustering techniques will be used for risk estimation." \\
"3. Return your output as a JSON object with three fields: 'relationships', 'clustering\_suggestions' and 'Results'."
}
}%
}
\vspace{10pt}

The model correctly identified the relationships in the dataset entities and suggested four Machine Learning clustering techniques which we briefly describe as follows:

\begin{enumerate}
    \item \textbf{Geospatial Clustering.} Performs geospatial risk analysis utilizing hashed geographical coordinates as features for the K-Means algorithm.
    To identify  high risk neighborhoods or regions where power theft is prevalent, for each produced cluster, the algorithm calculates the density of labeled samples relative to the total number of installations in that area.
    \item \textbf{Time Series Clustering.} Aggregates periodic data to form time series consumption profile for each subscriber. These profiles are then processed using the K-Means algorithm to form clusters based on pattern similarities. Subsequently, the algorithm merges the cluster assignments with the labeled samples to calculate risk for each cluster. Clusters exhibiting high density of labeled samples are categorized as high risk.
    \item \textbf{Mixed-Type Clustering.} The dataset comprises both numerical data, such as service capacity, and categorical features which include usage type and service category. Leveraging the K-Prototypes method, the algorithm groups similar installations and establishes baseline behaviors for specific facility types, like households and businesses. Finally, these clusters are cross-referenced with labeled samples to quantify risk.
    \item \textbf{Behavioral/Event Clustering.} Clustering based on the frequency and type of administrative events, e.g. disconnections and supplier changes. 
    The algorithm generates a behavioral profile for each subscriber and examines  whether high frequencies of specific requests correlates with higher risk groups.
\end{enumerate}

Subsequently, in the second stage, we initiated new session and the model was provided with the dataset description and the data report produced in the previous stage. For each suggested technique we prompted the model to generate code. In the first iteration, we noticed that the produced code did not utilize GPU to accelerate the procedures, which would cause unnecessary delays. Consequently, we had to refine the prompts for the suggested algorithms as follows:

\vspace{10pt}
\fbox{
\parbox{0.9\textwidth}{%
\texttt{
PROMPT: Based on the files uploaded generate code which will use GPUs if available for the suggested clustering technique. The code has to store the results in a file and clarify the distribution of the verified cases of non-technical losses in the generated clusters.
}
}%
}
\vspace{10pt}

Running the generated code in the third stage of the procedure revealed one critical issue. The second script exhibited high memory requirements. Consequently, we prompted the model for a memory-optimized version of the code, which eventually worked as expected.

Each script produced two files. The first contained the entities and information about the clusters into which they were classified. The second file contained statistics on the distribution of the entities and labeled samples within the clusters. 

\begin{figure}[H]
  \centering
  \begin{minipage}{0.5\textwidth}
    \centering
    \includegraphics[width=\textwidth]{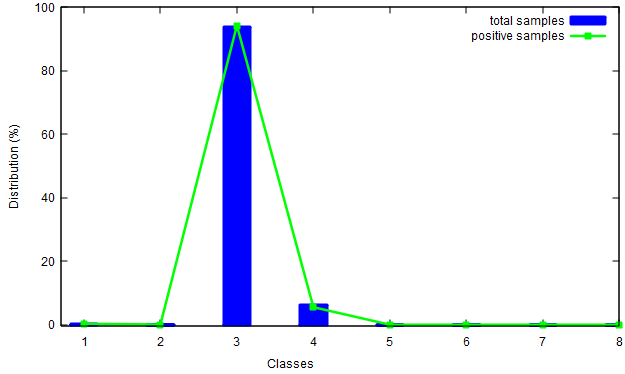}
    \caption{Results for first method.}
    \label{fig:image1}
  \end{minipage}%
\hfill
  \begin{minipage}{0.5\textwidth}
    \centering
    \includegraphics[width=\textwidth]{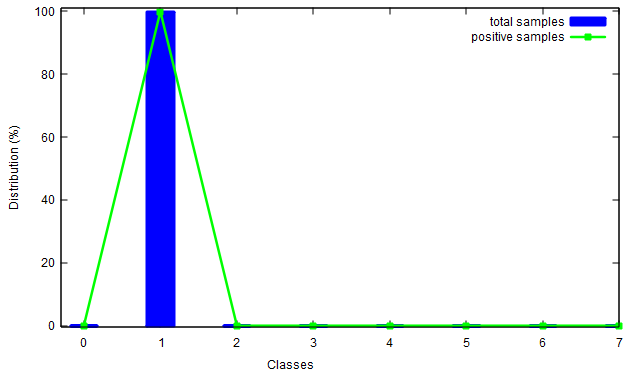}
    \caption{Results for second method.}
    \label{fig:image2}
  \end{minipage}
\end{figure}

\begin{figure}[H]
  \centering
  \begin{minipage}{0.5\textwidth}
    \centering
    \includegraphics[width=\textwidth]{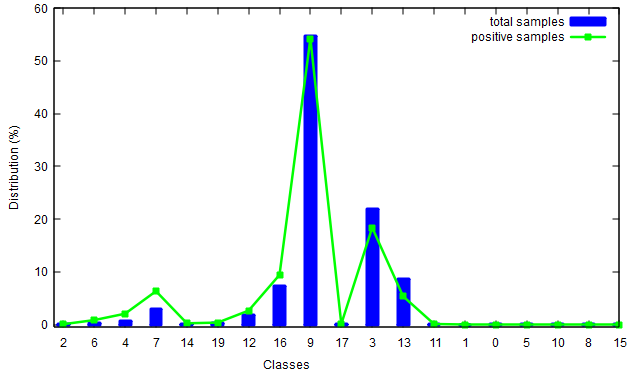}
    \caption{Results for third method.}
    \label{fig:image3}
  \end{minipage}%
  \hfill
  \begin{minipage}{0.5\textwidth}
    \centering
    \includegraphics[width=\textwidth]{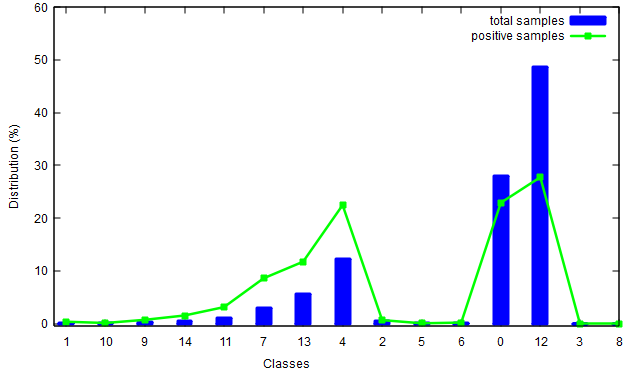}
    \caption{Results for fourth method.}
    \label{fig:image4}
  \end{minipage}
\end{figure}

We use these distributions to assess the outcome of the algorithms. Non-technical losses are expected to be a fraction of the total consumption, as such we expect the output of the procedure to include a cluster with a relatively small number of entities, which should contain a large portion of the labeled samples. This would effectively target a specific subset of the total dataset. Figures~\ref{fig:image1},~\ref{fig:image2},~\ref{fig:image3},~\ref{fig:image4} illustrate the distributions for the outputs of the four algorithms.

As the produced files containing the entity clustering were quite large, prompting the model to summarize and analyze the results in the fourth stage of the procedure was challenging. Furthermore, each clustering technique examined different features of the dataset and used different criteria to estimate the risky classes, as such we had to develop some algorithmic method to summarize these results. 

In the fourth stage, we provided the model with the dataset report and description, the descriptions of the clustering techniques and their respective results. Initially, we prompted the model to simply merge the results. The outcome was just a report based on the entities distribution, not an explicit report of risk estimation for each sample. As such, a refinement was necessary:

\vspace{10pt}
\fbox{
\parbox{0.9\textwidth}{%
\texttt{
PROMPT: For the whole set of the algorithms, I want a report that will summarize the results that they have generated and to merge the results in a csv indicating the risk estimation for each entity. \\
The produced files are [...]. Is there a script I need to run first? Generate the reports and any possible script you need me to run.
}
}%
}
\vspace{10pt}

The model identified that a script to scan the produced files was necessary, but the script it generated needed refinement. After a couple of iterations a subsequent prompt was formed as follows:

\vspace{10pt}
\fbox{
\parbox{0.9\textwidth}{%
\texttt{
PROMPT: Rate each installation according to the risk estimation in each algorithm result and estimate the final risk for each one based on the overall risk, entities with high risk in many results will have high final risk etc. \\
Add a column with the verified theft cases in every category.
}
}%
}
\vspace{10pt}

To address the request, the model proposed a consensus mechanism and generated the code that produced the final report.
It leverages a rank-based consensus voting mechanism which the model described as follows:

\vspace{10pt}
\fbox{
\parbox{0.9\textwidth}{%

\noindent
\texttt{Model output:}

\noindent
\textbf{1. Normalization:} Maps the risk percentage from each method to the specific installations.

\noindent
\textbf{2. Voting:} Flags an installation as "High Risk" (1 Vote) for a specific model if it falls into the top 20\% riskiest clusters of that model.

\noindent
\textbf{3. Ranking:} Prioritizes entities by the number of Machine Learning models that flagged them, followed by their average risk score.
}%
}
\vspace{10pt}

The model provided further clarifications on the mathematical formulation of the method as follows:

\vspace{10pt}
\fbox{
\parbox{0.9\textwidth}{%
\noindent
\texttt{Model output:}

\textbf{Mathematical Formulation.} Let $M$ be the set of $N$ learning models. For a given installation $i$, let $r_{i,m}$ be the risk probability estimated by model $m$.The binary vote $v_{i,m}$ is defined as:$$v_{i,m} = \begin{cases} 
1 & \text{if } r_{i,m} \geq P_{80}(R_m) \\
0 & \text{otherwise}
\end{cases}$$Where $P_{80}(R_m)$ is the $80^{th}$ percentile of all risk scores in model $m$. The Total Consensus Score ($C_i$) and Average Risk Intensity ($A_i$) are calculated as:$$C_i = \sum_{m=1}^{N} v_{i,m} \quad, \quad A_i = \frac{1}{N} \sum_{m=1}^{N} r_{i,m}$$ 

\textbf{Final Prioritization Logic.} The final ranking metric $S_i$ prioritizes the breadth of suspicion (Consensus) over the depth of suspicion (Intensity). This implies that an installation flagged by 4 different models is statistically more likely to be a true positive than an installation flagged extremely highly by only one model.$$S_i = C_i + \lambda \cdot A_i$$Where $\lambda$ is a scaling factor (set to $< 1/100$) ensuring that the Average Risk Intensity acts strictly as a tie-breaker between entities with equal Vote Counts.
}%
}
\vspace{10pt}

In this design, the largest number of votes, at most 4, signifies higher risk estimation, whereas samples without vote are categorized as low-risk or no-risk cases.

In the data fusion phase, the final consensus matrix integrates outputs from distinct feature views derived from separate utility datasets (i.e. installations, consumptions records and administrative requests). As such, the number of assessed entities exceeds the number of registered accounts.

For each class of entities, the script provides statistics of the associated cases of verified non technical losses to assess the clustering. Some cases of labeled samples, in particular 382, were not mapped to classified entities revealing the difficulty of data analysis in such datasets suffering from irregularity and sparsity.

\begin{figure}[htbp]
  \centering
  \includegraphics[width=0.7\columnwidth]{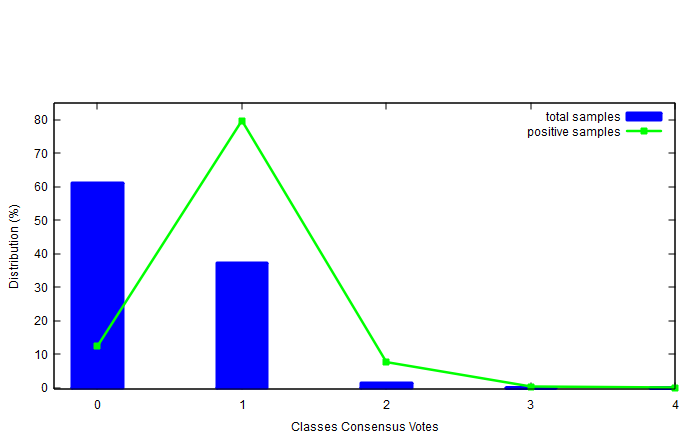}
  \caption{{Distribution of the samples in the five consensus classes.}}
  \label{fig:graph}
\end{figure}

The numerical results are presented in table~\ref{tab:result} and the distribution of samples across the five classes is illustrated in fig.~\ref{fig:graph}. Notably, 38.793\% of the total samples, in classes 1 to 4, are identified as risky entities and include 87.659\% of the labeled samples. Conversely, 61.207\% of the total samples are classified as low-risk or no-risk, in class 0. This distribution aligns with the expectation that the vast majority of consumers do not participate to non-technical losses.

\begin{table}[htbp]
  \centering
  \caption{{Consensus results}}
  \label{tab:result}
  \begin{tabular}{c|lll}
    \toprule
   \textbf{consensus votes} & \textbf{assessed entities} & \textbf{labeled entities} \\
    \midrule
    4  & 7 & ~1  \\
    3 & 491 & ~11  \\
    2 & 28576 & ~263  \\
    1 & 711886 & ~2758  \\
    0 & 1169089 & ~427  \\
    \bottomrule
  \end{tabular}
\end{table}

These results can be further refined to provide a more targeted analysis and generate a sharper distribution, thereby increasing the overall precision of the risk estimation. The model can also be guided to utilize more sophisticated techniques for improved efficiency. Our objective during this proof of concept was to implement the framework with the least possible intervention to demonstrate the robustness of our approach. Consequently, the prompts to the model were designed to be generic rather than restrictive and refinement requests were limited to those strictly necessary.

\section{Discussion}
\label{discussion}

The results presented in this work, are encouraging for the development of an autonomous data analysis and risk assessment system. Such a system could be based on agentic AI which constitutes a paradigm for the autonomous functionality of AI systems. Leveraging LLMs at their cores, AI agents can undertake the whole procedure of risk analysis, identifying required actions, selecting the necessary tools, and finally using them to achieve the objective. 

By design, agents appear capable of accomplishing the objectives of the proposed framework. However, limitations in the current state of the art, identified during the proof of concept, coupled with inherent privacy issues, raise concerns regarding the efficacy and safety of this design.

LLMs, and by extension agentic AI, although impressively useful are still based on immature technologies. Foundation models have not yet managed to escape from hallucination phenomena and flaws in generated outputs. Furthermore, the AI alignment problem persists. This limits their credibility and necessitates human supervision of both the running procedures and the generated results. Within this context, full autonomy is risky and a guided approach with Human-in-the-Loop appears more rational.

The use of LLMs provided as-a-service raises privacy issues. Unless a model is installed on-premises, inaccessible from the outside world, users can never be sure how the sensitive data they input is used. Advanced models require sophisticated high-end equipment and usually the most advanced models are not open source. Consequently, relying solely on on-premises infrastructure is challenging. In many cases it is necessary to use LLMs as-a-service, raising questions about data privacy.

\bibliography{sn-bibliography}

\end{document}